\pdfoutput=1

\documentclass[11pt]{article}

\usepackage{acl}

\usepackage{times}
\usepackage{latexsym}

\usepackage[T1]{fontenc}

\usepackage[utf8]{inputenc}

\usepackage{microtype}

\usepackage{hyperref}
\usepackage{url}

\usepackage{comment}
\usepackage{tabularx}
\usepackage{booktabs}
\usepackage{multirow}
\usepackage{colortbl}
\usepackage{xcolor}
\usepackage{amsfonts}
\usepackage{bbm}
\usepackage{wrapfig}
\usepackage{tabularx}
\usepackage{hyperref}

\usepackage{caption}
\usepackage{subcaption}
\usepackage{graphicx}
\usepackage{subfiles}

\newcommand\ti[1]{\textit{#1}}

\newcommand\tf[1]{\textbf{#1}}
\newcommand\ttt[1]{\texttt{#1}}

\mathchardef\mhyphen="2D

\usepackage{amsmath,amsfonts,bm}

\def\1{\bm{1}}

\makeatletter
\newcommand{\ve}{\@ifnextchar\bgroup{\velong}{{\bm{e}}}}
\newcommand{\velong}[1]{{\bm{#1}}}
\makeatother

\def\ve{{\mathbf{e}}}

\DeclareMathAlphabet{\mathsfit}{\encodingdefault}{\sfdefault}{m}{sl}
\SetMathAlphabet{\mathsfit}{bold}{\encodingdefault}{\sfdefault}{bx}{n}

\def\gD{{\mathcal{D}}}

\def\gV{{\mathcal{V}}}

\def\gX{{\mathcal{X}}}

\DeclareMathOperator*{\argmax}{arg\,max}

\title{Measuring Inductive Biases of In-Context Learning \\ with Underspecified Demonstrations}

\author{Chenglei Si$^{1*}$ \hspace{0.2cm} Dan Friedman$^{2*}$   \\
 \textbf{Nitish Joshi}$^{3}$ \hspace{0.1cm}  \textbf{Shi Feng}$^{4}$ \hspace{0.1cm}  \textbf{Danqi Chen}$^{2}$ \hspace{0.1cm}   \textbf{He He}$^{3}$\\
  $^{1}$University of Maryland \hspace{0.5cm}
  $^{2}$Princeton University \\
  $^{3}$New York University \hspace{0.6cm}
  $^{4}$University of Chicago \\
  \texttt{clsi@umd.edu}, \hspace{0.1cm} \texttt{dfriedman@cs.princeton.edu} \\
}

\begin{document}
\maketitle
\renewcommand{\thefootnote}{\fnsymbol{footnote}}
\footnotetext[1]{Equal contribution.}
\renewcommand{\thefootnote}{\arabic{footnote}}



\begin{abstract}
In-context learning (ICL) is an important paradigm for adapting large language models (LLMs) to new tasks, but the generalization behavior of ICL remains poorly understood.
We investigate the inductive biases of ICL from the perspective of feature bias: which feature ICL is more likely to use given a set of \ti{underspecified} demonstrations in which two features are equally predictive of the labels. 
First, we characterize the feature biases of GPT-3 models by constructing underspecified demonstrations from a range of NLP datasets and feature combinations.
We find that LLMs exhibit clear feature biases---for example, demonstrating a strong bias to predict labels according to sentiment rather than shallow lexical features, like punctuation.
Second, we evaluate the effect of different interventions that are designed to impose an inductive bias in favor of a particular feature, such as adding a natural language instruction or using semantically relevant label words.
We find that, while many interventions can influence the learner to prefer a particular feature, it can be difficult to overcome strong prior biases. 
 Overall, our results provide a broader picture of the types of features that ICL may be more likely to exploit and how to impose inductive biases that are better aligned with the intended task.\footnote{Our code and data are publicly available at \\ \href{https://github.com/NoviScl/AmbigPrompt}{https://github.com/NoviScl/AmbigPrompt}.}

\end{abstract}

\begin{figure}[ht!]
\centering 
  \includegraphics[trim={0 0.1cm 0.1cm 0}, scale=0.53]{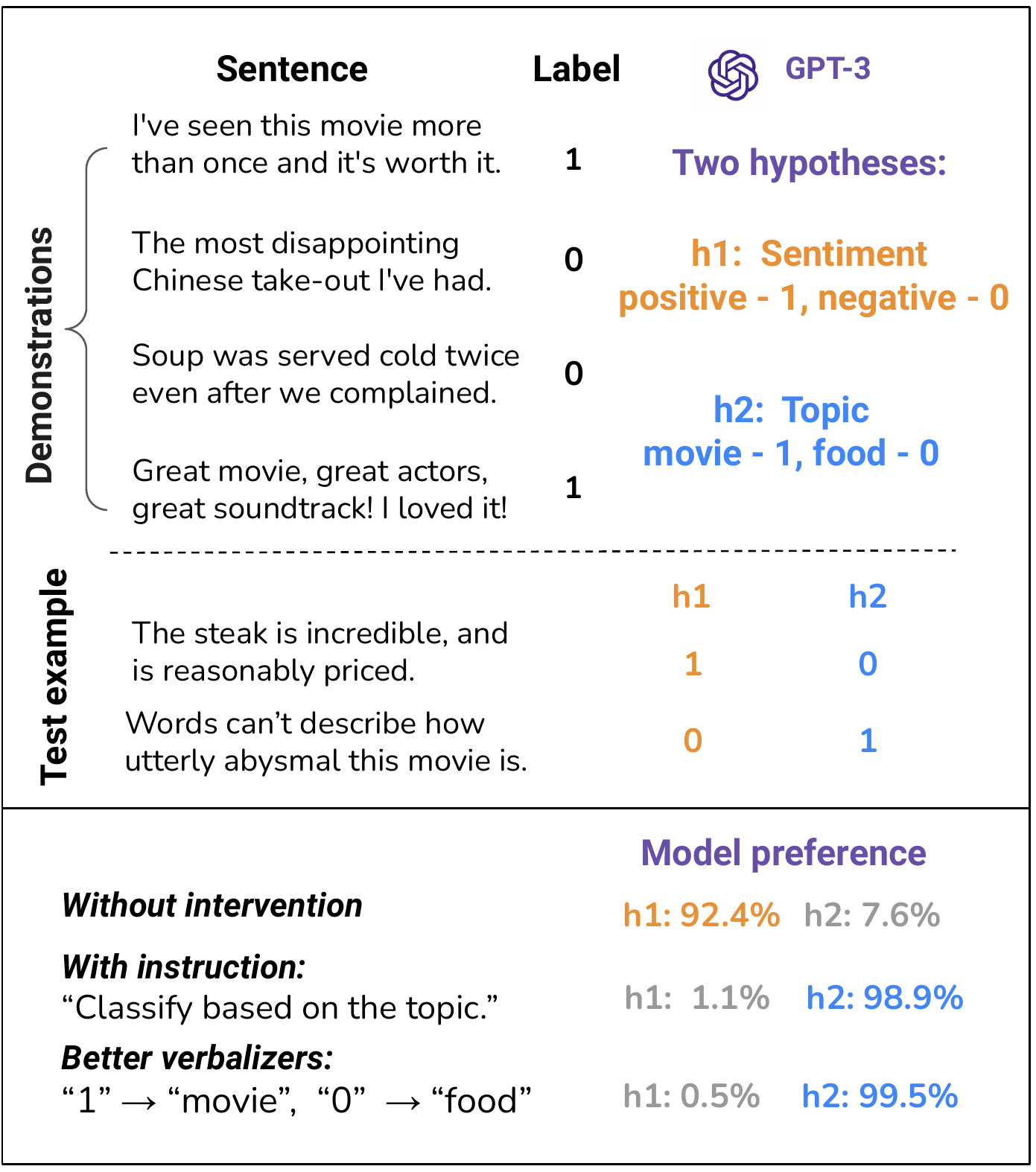}
  \caption{
We prompt language models with underspecified demonstrations in which two features are equally predictive of the label.
In this case, the decision rule could be based on either sentiment (positive vs. negative) or topic (movie vs. food).
We measure feature biases by testing the model on disambiguation examples where the two hypotheses disagree, such as positive restaurant reviews.
Here, \textsc{GPT-3} strongly favors the sentiment hypothesis, but we can encourage it to prefer the topic feature using various interventions, e.g., adding a natural-language instruction or using verbalized labels.
  }
  \label{fig:teaser}
\end{figure}

\section{Introduction}

In-context learning (ICL) is an increasingly popular paradigm for adapting large language models (LLMs) to downstream tasks~\citep{Brown2020LanguageMA}. It works by prompting LLMs with a small set of examples that demonstrate the input and output of a task, without requiring any parameter updates.
However, a key limitation of ICL is that it can only incorporate a small number of demonstration examples (e.g., 16) due to the context length limit of Transformer models, meaning that most tasks will be highly underspecified.
For example, as shown in Figure~\ref{fig:teaser}, we present the model with an underspecified text classification problem. The sentences with label `1' are positive reviews of movies, and the sentences with label `0' are negative reviews of restaurants.
From the demonstrations, it is unclear whether the labels are determined by the sentiment feature (positive vs. negative) or the topic feature (movie vs. food).
Moreover, due to the limited context window, it is difficult to specify the task by supplying a large number of additional training examples.
Instead, ICL can succeed only if (a) the LLM has an inductive bias that happens to align well with the given task or (b) there is a mechanism for imposing an inductive bias on the system, which can specify the task (e.g., whether it is sentiment classification or topic classification) without incorporating more training examples.

In this paper, we study the inductive biases of ICL with LLMs and measure the effectiveness of different interventions to steer ICL towards a particular hypothesis.
We focus on feature biases: a tendency to use one feature rather than another given a prompt in which the two features predict the label equally well.
As illustrated in Figure~\ref{fig:teaser},
by evaluating the model on sentences where the two features disagree---such as positive restaurant reviews---we can measure the feature bias of the learning algorithm, and we can attempt to modify the feature biases through various interventions applied to the prompt.

In the first part of the paper, we characterize the feature biases of ICL by constructing underspecified demonstrations from a variety of NLP datasets and feature pairs.
We find that ICL models exhibit some clear feature biases. 
For example, in a sentiment analysis setting, the LLMs we study exhibit a strong tendency to generalize on the basis of sentiment rather than other equally predictive features, including topics of sentences.
On sentence-pair tasks, such as question answering and natural language inference, the GPT-3 model~\cite{Brown2020LanguageMA} generally prefers shallow lexical features, while the instruction-tuned models~\cite{Ouyang2022TrainingLM} generalize more consistently with the labels associated with those datasets.
Such feature biases could potentially be problematic---users could have intended either of the two predictive features as the actual task. When the model's feature bias does not align with the intended task, we want the model to be steerable with appropriate interventions.

In the second part of the paper, we measure whether simple modifications of the prompt can supply an inductive bias to the ICL learning algorithm, steering the model to generalize according to one feature rather than another.
These interventions include using natural-language instructions or explanations and using label words that are semantically related to the intended feature.
As a baseline, we compare these methods with unambiguous prompts, in which some demonstration examples are consistent with one hypothesis but not the other.
We find that these interventions are most effective when the model does not have a strong feature bias, or already has a bias in favor of the intended
task feature.
They are less effective at steering the model to use one feature when it has a strong bias to use another feature. 
For example, the instruction-tuned model generalizes on the basis of sentiment despite adding instructions and even disambiguating evidence in favor of lexical features like punctuation.
Interestingly, we find that data-independent methods, like using semantically relevant label words, sometimes have a stronger effect than providing unambiguous data.

These results underscore some of the challenges involved with using ICL as a general-purpose machine learning method, complementing a growing body of work that has attempted to explain how ICL works from an empirical~\citep[e.g.,][]{min2022rethinking,webson2022prompt,chan2022transformers} and theoretical~\citep{xie2022an,akyurek2022learning,von2022transformers} point of view.
On one hand, the strong inductive biases of LLMs are helpful when they happen to be well aligned with the given task, enabling ICL to generalize successfully from very few training examples.
Moreover, simple modifications to the prompt are often successful at steering the model towards a particular feature in underspecified settings. 
On the other hand, simple prompting methods cannot systematically align the model with user intention: they have limited effectiveness when the model's feature biases conflict with the intended task. 

\section{Setup}

\subsection{Measuring Feature Biases}
We consider text classification problems, where $x \in \gX$ is a text input and $h_1, h_2 : \gX \to \{0, 1\}$
are two binary feature functions.
For example, $h_1$ may be a sentiment classifier (returning 0 if $x$ is negative and 1 if it is positive), and $h_2$ is a domain classifier indicating whether $x$ is a review of a movie or a restaurant.
We consider a learning algorithm $\ell$, defined as a mapping from a dataset $\gD \subseteq \gX \times \{0, 1\}$ to a classifier $f: \gX \to \{0, 1\}$.
Given a learning algorithm $\ell$ and a pair of feature functions $h_1, h_2$, our aim is to understand whether $\ell$ tends to return classifiers more similar to $h_1$ or $h_2$,
given that $h_1$ and $h_2$ have the same accuracy on $\gD$.

In the context of ICL, we measure this property behaviorally by prompting language models with a set of 
\textbf{underspecified demonstrations} $\gD_{\text{demo}}$ and evaluating the resulting function $f = \ell(\gD_{\text{demo}})$ 
on a \textbf{disambiguating dataset} $\gD_{\text{test}}$.
The underspecified demonstrations are examples $\gD_{\text{demo}} \in \gX \times \{0, 1\}$ such that, for all $(x, y) \in \gD_{\text{demo}}$, $y = h_1(x) = h_2(x)$; and we ensure that the labels are balanced on $\gD$.
The disambiguating dataset $\gD_{\text{test}}$ is constructed so that, for all $x$, $h_1(x) \neq h_2(x)$, and the dataset is balanced with respect to $h_1(x)$ and $h_2(x)$.
A simple example is illustrated in Figure~\ref{fig:teaser}.

We measure whether $f$ is more consistent with $h_1$ or $h_2$ by comparing the predictions of $f$, $h_1$, and $h_2$ on $\gD_{\text{test}}$.
For a given feature function $h$, we define the \textbf{accuracy on $h$} as the proportion of examples for which $f(x) = h(x)$: 
\[
h\mathrm{\mhyphen accuracy} = \frac{1}{|\gD_{\text{test}}|} \sum_{x \in \gD_{\text{test}}} \mathbbm{1}[f(x) = h(x)].
\]
The difference between $h_1$-accuracy and $h_2$-accuracy on $\gD_{\text{test}}$ can be interpreted as a feature bias: for example, a high value of $h_1$-accuracy indicates that the model is more likely to predict the same label as $h_1$ in situations where $h_1$ and $h_2$ disagree.
$h_1$-accuracy and $h_2$-accuracy always add up to 1 and, because $\gD_{\text{test}}$ is balanced, an $h_1$-accuracy of 0.5 indicates that the model does not exhibit a strong bias towards either feature.

\subsection{In-Context Learning}
\label{sec:in_context_learning}

The learning algorithms we consider in this paper are based on in-context learning (ICL) of LLMs~\citep{Brown2020LanguageMA}.
A language model $p_{\theta}(w)$ assigns scores to strings $w \in \gV^*$, where $\gV$ is a discrete vocabulary.
The input to ICL is a language model $p_\theta(w)$ and a function that converts a dataset $\gD$ and a single test example $x_{\text{test}}$ into a prompt $c(\gD, x_{\text{test}}) \in \gV^*$.
We consider the basic form of ICL, which consists of\footnote{Additionally, we can incorporate natural-language instructions and explanations, as we will discuss in Section~\ref{sec:interventions}.}:
(1) an \textbf{instance template} $t: \gX \to \gV^*$ that encodes each data instance $x$ as a string;  
(2) a \textbf{label verbalizer} $v: \{0, 1\} \to \gV^*$ that encodes each label as a string.
For the first part of our analysis on measuring feature biases (Section~\ref{sec:experiments}), we adopt the simplest format and define the instance template as $t(x) = \text{``Input: \ttt{\$x} Label: ''}$, and the label verbalizer as $v(0) = \text{``0''}$ and $v(1) = \text{``1''}$.
The prompt $c$ is then the concatenation of 
 $t(x_i)$ and 
 $v(y_i)$  for all demonstration examples $(x_i, y_i) \in \gD$; and lastly the test instance $t(x_{\text{test}})$.
The resulting classifier is:
\begin{align*}
f(x_{\text{test}}; \gD) = \argmax_y p_{\theta}(v(y) \mid c(\gD, x_{\text{test}})).
\end{align*}
ICL is known to be sensitive to the order of the demonstrations~\citep{lu2022fantastically} and to demonstrate other biases that are orthogonal to this study, including majority label bias and recency bias~\citep{zhao2021calibrate}.
We control for these by ordering the demonstration examples randomly 
and performing label calibration, following~\cite{zhao2021calibrate}.

\section{Data Construction}
\label{sec:datasets}
\begin{table*}[ht!]
\small
\begin{center}
\begin{tabular}{l l l}
\toprule
\textbf{Task} & \textbf{Dataset} & \textbf{Hypotheses} \\
\midrule
\textbf{Single-sentence classification} \\
Sentiment analysis & IMDb + Yelp & \tf{Sentiment (positive vs. negative)} \\
& & Domain (IMDb vs. Yelp) \\
& & Length (short vs. long) \\
& & Terminal punctuation (exclamation vs. period) \\
& & Contains word (``nice''/``food'') \\
& & Capitalization (lowercase vs. uppercase) \\
Toxicity classification & CivilComments & \tf{Toxicity (toxic vs. non-toxic)} \\
& & Gender (female vs. male) \\
& & Sexuality (LGBTQ vs. non-LGBTQ) \\
& & Religion (Muslim vs. Christian; Muslim vs. Jewish) \\
& & Race (Black vs. White; Asian vs. White) \\
& & Length (short vs. long) \\
& & Capitalization (lowercase vs. uppercase) \\
\midrule
\textbf{Sentence-pair classification} \\
Natural language inference & MultiNLI & \tf{Entailment (entailment vs. non-entailment)} \\
& & Domain (government vs. fiction; government vs. telephone) \\
& & Lexical overlap (overlap vs. non-overlap) \\
& & Hypothesis length (long vs. short) \\
& & Hypothesis negation (contains ``not'', ``n't'', ``no'') \\
Question answering  & BoolQ & \tf{Answer (yes vs. no)} \\
& & Question word (``is/was'' vs. ``do/does/did'') \\
& & Lexical overlap (overlap vs. non-overlap) \\
& & Question structure (``is x the same as y'') \\
& & Passage length (short vs long) \\
\bottomrule
\end{tabular}
 \caption{The datasets and features we study in this paper. We treat the first feature for each task as the default feature (referred to as $h_1$) and the others as distractors ($h_2$). We measure feature biases between the default feature (in bold) and each of the distractor features.}
 \label{tab:datasets}
\end{center}
\end{table*}

We choose datasets to cover four different NLP tasks, including both single-sentence and sentence-pair classification. 
For sentiment analysis, we use 
 IMDb~\citep{maas2011learning} and Yelp~\citep{asghar2016yelp} datasets; 
 for toxicity classification, we use the CivilComments dataset~\citep{borkan2019nuanced}; 
 for natural language inference, we use the MNLI dataset~\citep{williams2018broad}; and for question answering, we use BoolQ~\citep{clark2019boolq}.

For each dataset, we select the original classification label as the default feature and denote it as $h_1$. 
We select alternative comparison features ($h_2$) using existing metadata or simple, deterministic functions, chosen to reflect realistic sources of ambiguity or spurious correlation that have been studied in prior work~\cite{HANS,Gururangan2018AnnotationAI,poliak2018hypothesis,Joshi2022AreAS}, as well as common shallow features such as length, capitalization, and the presence of particular words or punctuation marks. 
All datasets and features we use are listed in Table~\ref{tab:datasets}, which we elaborate below:
%

(1) For sentiment analysis, the default feature is the sentiment, and the alternative features include: domain or source of the review (based on whether it is from IMDb or Yelp), length of the review (longer or shorter than a threshold), the final punctuation mark of the review (exclamation mark or period), whether it contains certain keywords (``food'' and ``nice''), and whether it contains uppercase words (e.g., ``THIS''). 

(2) For toxicity classification, the default feature is whether the comment is toxic. The alternative features are: gender, sexuality, religion, and race mentioned in the comment (all based on human-annotated meta-data), and its length and capitalization (whether containing uppercase words).
%

(3) For NLI, the default feature is the entailment relationship between the sentence pair, and we condense the neutral and contradiction classes as non-entailment to cast the task as binary classification. For alternative features, we consider: domain of the text (from the genre meta-data); lexical overlap, following the definition in HANS~\citep{HANS}; whether the hypothesis is shorter or longer than the premise; and whether the hypothesis contains negation words (``no'', ``not'', ``n't''). 

(4) For question answering, the default feature is whether the answer is yes or no, and the alternative features are: the question word, whether all words from the question also appear in the passage (lexical overlap), question structure (whether it is a comparison question), and passage length. 
These features are potential spurious features in QA that have been documented in prior work~\cite{Pezeshkpour2021CombiningFA,Shinoda2022WhichSS}.


\section{Measuring Feature Biases}
\label{sec:experiments}

\begin{figure*}[ht]
    \centering
     \begin{subfigure}[b]{\textwidth}
         \centering
         \includegraphics[width=0.08\textwidth]{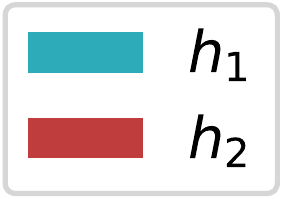}
     \end{subfigure}
     \hfill
     \begin{subfigure}[b]{0.49\textwidth}
         \centering
         \includegraphics[width=\textwidth]{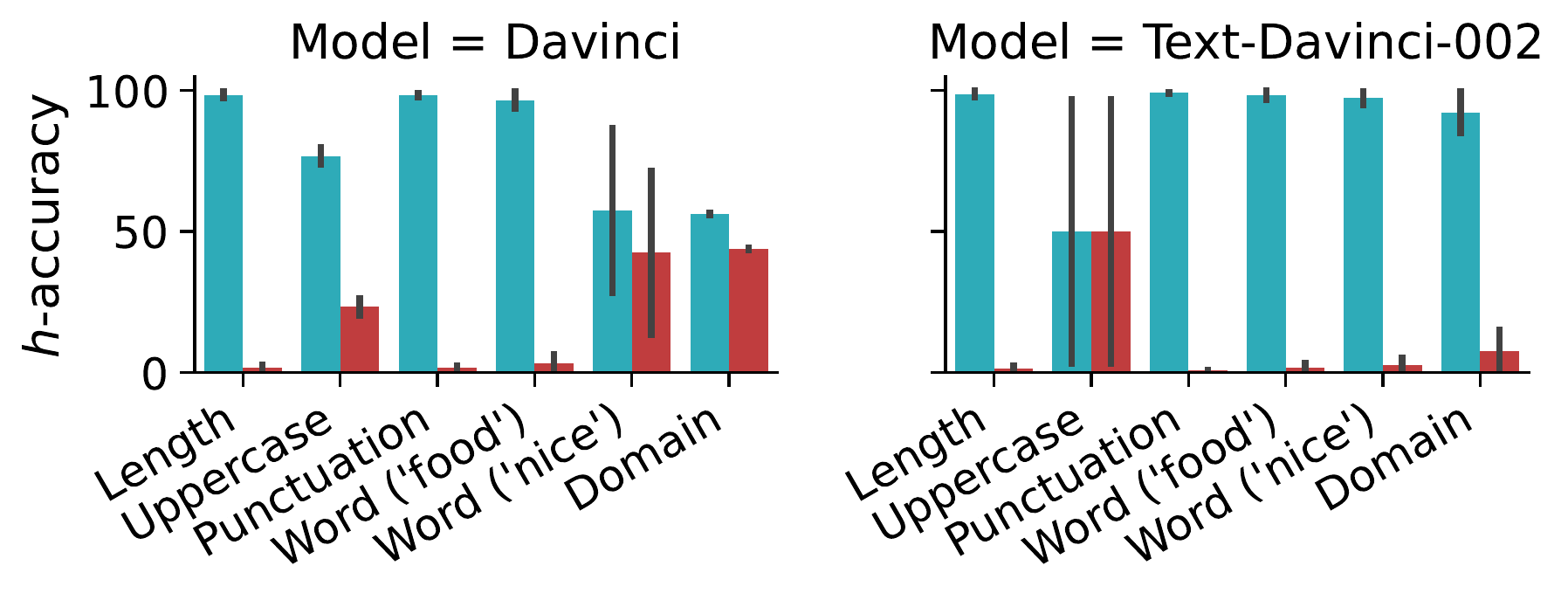}
         \caption{Sentiment analysis}
         \label{fig:sentiment}
     \end{subfigure}
     \hfill
     \begin{subfigure}[b]{0.49\textwidth}
         \centering
         \includegraphics[width=\textwidth]{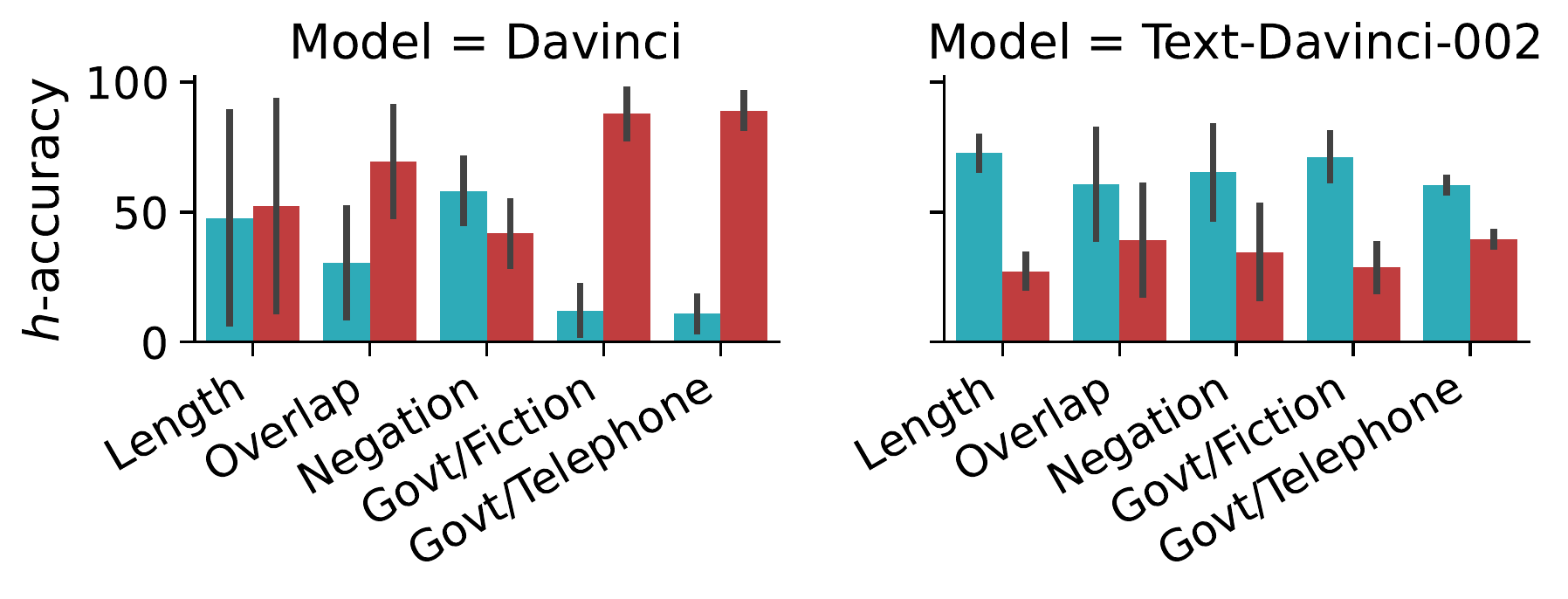}
         \caption{MultiNLI}
         \label{fig:mnli}
     \end{subfigure}
     \hfill
     \begin{subfigure}[b]{0.49\textwidth}
         \centering
         \vspace{0.2em}
         \includegraphics[width=\textwidth]{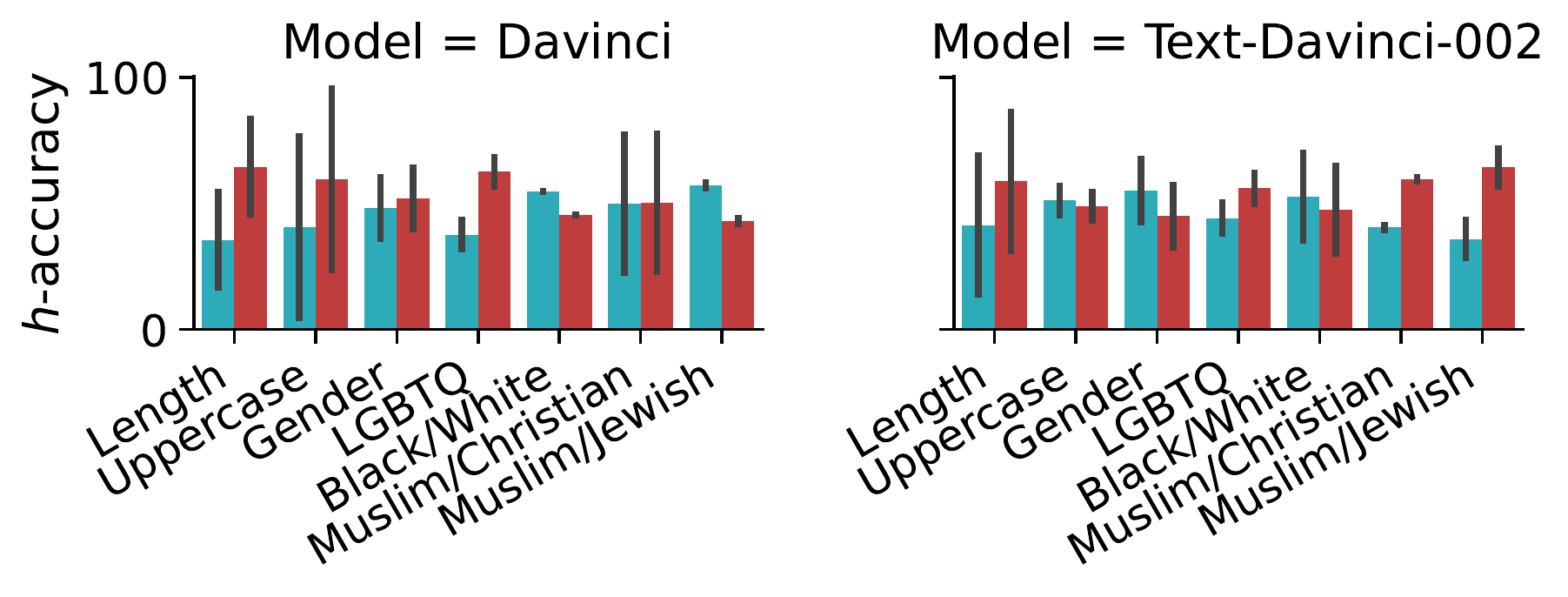}
         \caption{CivilComments}
         \label{fig:civil_comments}
     \end{subfigure}
     \hfill
     \begin{subfigure}[b]{0.49\textwidth}
         \centering
         \vspace{0.2em}
         \includegraphics[width=\textwidth]{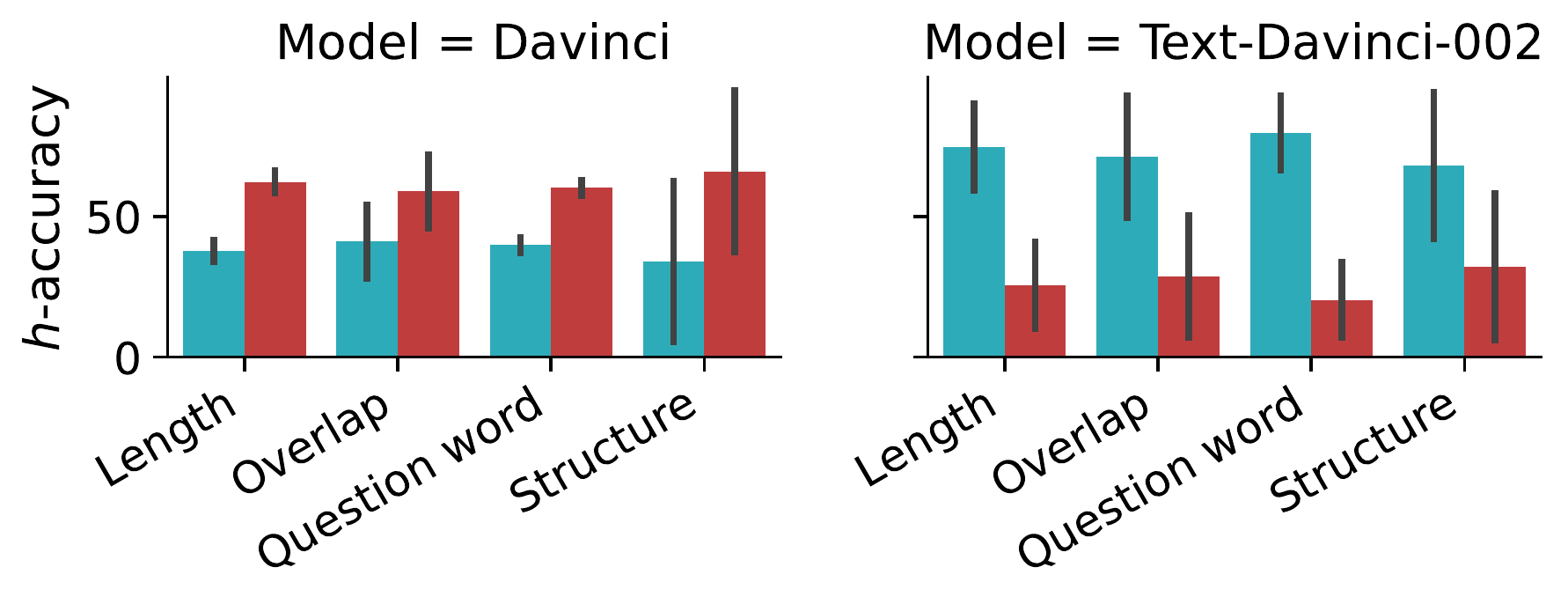}
         \caption{BoolQ}
         \label{fig:boolq}
     \end{subfigure}
    \caption{
The generalization behavior of two LLMs given demonstration examples that could support both $h_1$ (sentiment, entailment, toxicity, and question-answer) and an alternative feature $h_2$.
The y-axis is the accuracy on a balanced set of disambiguating examples, where $h_1(x) \neq h_2(x)$, measuring accuracy using either $h_1(x)$ or $h_2(x)$ as the ground truth label.
Both models generally show a strong preference for the sentiment feature.
On the sentence-pair datasets (MultiNLI and BoolQ),
the Davinci model tends to prefer the distractor feature, while the Text-Davinci-002 model, which was trained with human feedback, always prefers the original task hypothesis. 
Error bars represent the standard deviation over three random seeds, which correspond to three sets of randomly sampled demonstration examples as prompts.  
}    \label{fig:results_base}
\end{figure*}

\subsection{Experiment Details}
\label{sec:experiment_details}

\paragraph{Evaluation protocol.}
For all experiments, we randomly sample a balanced set of 16 demonstration examples (randomly shuffled) to form the prompt.
For eight of the examples, $y = h_1(x) = h_2(x) = 0$; for the other eight, $y = h_1(x) = h_2(x) = 1$.
For each experiment, we randomly sample three sets of prompts and report the average performance on a set of 1,200 test examples, balanced between examples with $h_1(x) = 0/h_2(x) = 1$ and $h_1(x) = 1/h_2(x) = 0$.
In this baseline setting, the instance template is $t(x) = \text{``Input: \ttt{\$x} Label: ''}$, and the label verbalizer provides no additional information about the task: $v(0) = \text{``0''}$ and $v(1) = \text{``1''}$. 

\paragraph{Models.} We focus on the \textsc{Text-Davinci-002} and \textsc{Davinci} checkpoints of GPT-3 mainly because smaller-scale models often perform no better than random on tasks like NLI.
The main differences between the two checkpoints are the pretraining data and the additional instruction tuning for \textsc{Text-Davinci-002}~\citep{Ouyang2022TrainingLM}.
For all experiments, we use a temperature value of 0 in GPT-3 decoding, and all experiments involving the OpenAI API were conducted in January 2023.

\paragraph{Metric.} We report the $h$-accuracy for each feature. Higher $h$-accuracy indicates a higher preference for that feature. We denote the default feature $h_1$ (sentiment, toxicity, entailment, answer) and the alternative
features $h_2$. 

\subsection{Results}
\label{sec:measuring_inductive_biases}

We present ICL results on all the datasets and pairs of features in Figure~\ref{fig:results_base}, and note several interesting trends as follows: 

\paragraph{LLMs often have clear feature biases.}
For example, in the sentiment analysis setting (Figure~\ref{fig:sentiment}), both models generally show a strong preference to predict labels according to the sentiment of the sentence rather than other features, such as sentence length or the presence of individual lexical items like the word ``food''. Such biases can be helpful when they are aligned with the intended task. 

On the other hand, we do not observe clear feature preferences in the CivilComments dataset (Figure~\ref{fig:civil_comments}), suggesting that these models may not have a strong feature biases in this setting.

\paragraph{The instruction-tuned model is generally more aligned with standard dataset labels.}
While both \textsc{Davinci} and \textsc{Text-Davinci-002} show similar feature biases in the sentiment analysis setting, they show very different biases on the the sentence pair datasets MultiNLI (Figure~\ref{fig:mnli}) and BoolQ (Figure~\ref{fig:boolq}):
the \textsc{Davinci} model tends to prefer the shallow distractor features, such as lexical overlap, while \textsc{Text-Davinci-002} tends to prefer the semantic feature associated with the dataset---either the entailment relationship or the answer to the question.
This may be due to some aspect of instruction tuning, which might have exposed the model to similar tasks.


\section{Comparing Interventions}
\label{sec:interventions}
Our findings that LLMs can have strong feature biases have important implications: when the LLMs' biases do not align with users' intended task, such biases would hurt performance. To resolve such misalignment, 
we explore a set of intervention methods designed to encourage the model to prefer one feature over another, examining whether the $h$-accuracy for the intended feature indeed increases.

\subsection{Experiment Details}
\label{sec:intervention_details}

We now  consider more variants of ICL that can be decomposed into four components that are commonly used in various prompting methods. In addition to the instance template $t$ and label verbalizer $v$ described in Section~\ref{sec:in_context_learning}, we also introduce:
(1) An \textbf{instruction} $s \in \gV^*$, which is prepended to the prompt;
and
(2) a free-form \textbf{explanation} $e: \gX \to \gV^*$ after each input text $t(x)$ and before the label $v(x)$. 
The prompt $c$ is then the concatenation of 
$s$, followed by 
 $t(x_i)$;
$e(x_i)$; 
 $v(y_i)$  for all demonstration examples $(x_i, y_i) \in \gD$; and lastly the test instance $t(x_{\text{test}})$.

Each intervention operates on a combination of the above components.
We apply these interventions separately and compare  with the baseline rather than applying all interventions on top of each other in order to analyze the impact of each of the methods. 
We compare interventions designed to steer the model towards $h_1$ and $h_2$ as the intended feature respectively, to understand the effectiveness of interventions towards different features.


\begin{itemize}
\item Recall that in the \textbf{baseline} setting, there is no instruction or explanation ($s$ and $e$ are empty strings). We simply concatenate demonstration examples as the prompt, and use ``1'' and ``0'' as the verbalizer. 

\item In the \textbf{semantic verbalizer} setting, the verbalizer selects label words that are semantically related to a chosen feature in order to hint at the intended task. For example, if the intended feature is \textit{sentiment}, then $v(0) = \text{``negative''}$ and $v(1) = \text{``positive''}$; and if the intended feature is \textit{length}, then $v(0) = \text{``short''}$ and $v(1) = \text{``long''}$. Our choice of verbalizers is inspired by prior work~\cite{Gao2021MakingPL,Shi2022kNNPromptNN} and we list all of them in Table~\ref{tab:verbalizers}. 

\item In the \textbf{instruction} setting, we add a prefix string describing the intended feature and instructing the model to use this feature. We format our instructions mostly following prior work such as Natural Instructions~\citep{Mishra2021NaturalIB,Wang2022SuperNaturalInstructionsGV}, and we list all our instructions in Tables~\ref{tab:instructions_1} and ~\ref{tab:instructions_2}. 

\item In the \textbf{explanation} setting, we append a template explanation after each demo example to explain why the prediction is made 
based on the intended feature,
 formatted in a similar manner as Chain-of-Thought prompting~\cite{Wei2022ChainOT} and ``explain-then-predict''~\citep{Ye2022TheUO}.
Since hand-written explanations are hard to obtain, we create fixed human-written templates for each feature value.
For example, for the punctuation feature, all positive examples have the explanation ``The review ends with an exclamation mark. Therefore, the answer is 1''.
We list all our template explanations in Tables~\ref{tab:explanations_1} and ~\ref{tab:explanations_2}.

\item Finally, we include a \textbf{disambiguation} setting, in which we change half of the demonstration examples to those that disambiguate the task in favor of the intended feature. For example, to steer the model towards $h_1$, the demonstration includes examples such that $h_1(x) \neq h_2(x)$ and $y = h_1(x)$. Intuitively, this provides additional evidence for the model to differentiate the intended feature. 

\end{itemize}
We measure the effectiveness of the intervention in terms of the increase in $h$-accuracy relative to the baseline model, where $h$ is the intended
feature.

\begin{table}[t!]
    \centering
    \small
    \setlength{\tabcolsep}{5pt}{
\begin{tabular}{lrr|rr}
\toprule
& \multicolumn{2}{c|}{\textit{Steer towards $h_1$}} & 
\multicolumn{2}{c}{\textit{Steer towards $h_2$}} \\
 \textit{Intervention} &  \textbf{Davinci} &  \textbf{TD002} &  \textbf{Davinci}  &  \textbf{TD002}  \\
\midrule
\ti{Baseline}         &     \ti{39.5} &              \ti{59.1} &          \ti{46.9} &                   \ti{30.5} \\
\midrule
Verbalizer   &     +11.9 &              +7.1 &          +15.6 &                   +24.4 \\
Instruction  &     +1.6 &              +12.2 &          -2.4 &                   +24.2 \\
Explanation  &     +14.4 &              +6.9 &          +14.3 &                   +33.8 \\
\midrule
Disambig     &     +12.9 &              +9.4 &          +18.6 &                   +21.1 \\
\bottomrule
\end{tabular}
    }
\caption{
The impact of different intervention strategies on $h_1$-accuracy (left) or $h_2$-accuracy (right), averaged over features and datasets. (The \textit{Steer towards $h_1$} experiments exclude the sentiment analysis datasets, because the models already strongly prefer $h_1$ even without interventions.)
We report the change in accuracy relative to the baseline.
Higher values indicate that the intervention is more effective at steering the model to predict labels according to the given feature.
TD002: \textsc{Text-Davinci-002}.
}
\label{tab:interventions_aggregated}
\end{table}

\begin{figure*}[ht]
    \centering
    \includegraphics[width=\textwidth]{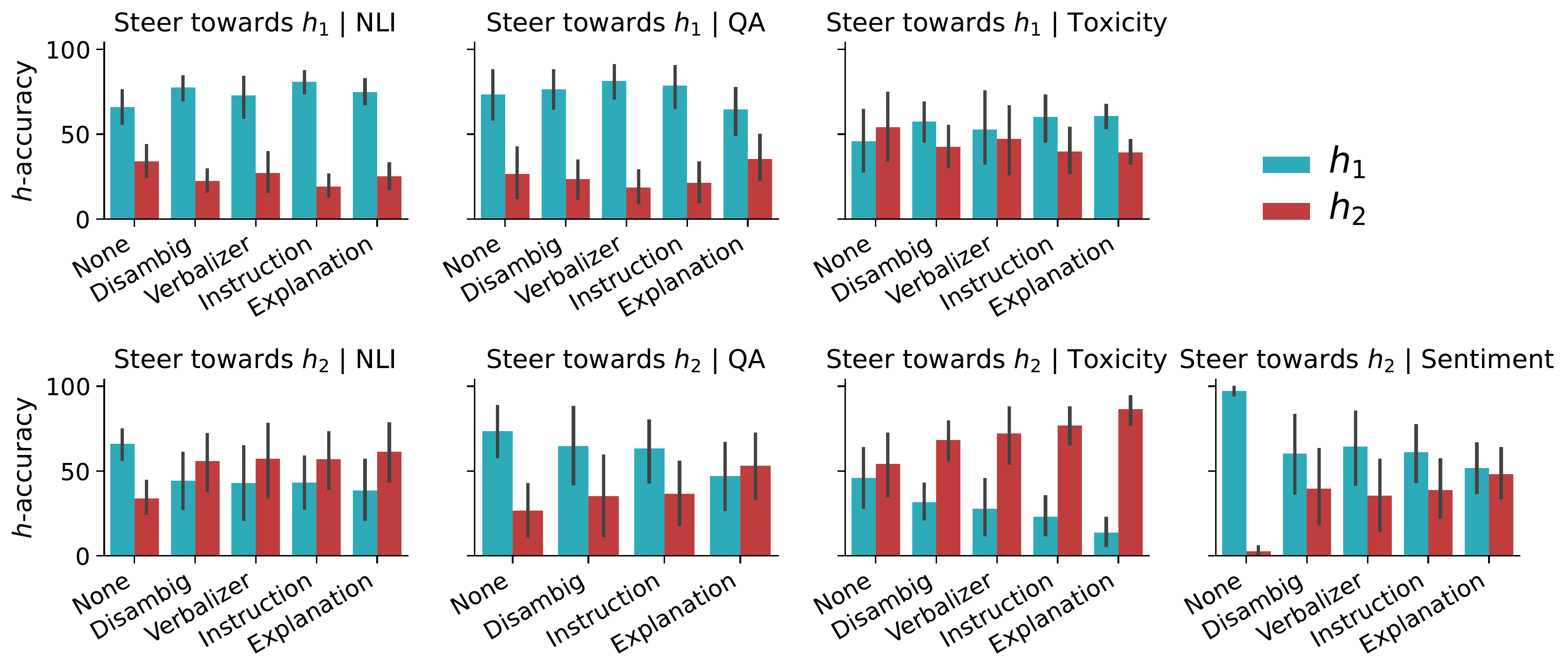}
    \caption{
    Generalization results of \textsc{Text-Davinci-002} using different interventions, aggregated over feature pairs.
    In the first row, the intervention is designed to steer the model towards $h_1$, so we expect an increase in the $h_1$-accuracy (the blue bars) compared to the baselines (the first pair of bars in each plot).
    In the second row, the intervention is designed to steer the model towards $h_2$, so we expect an increase in the $h_2$-accuracy (red bars).
    The interventions are most successful when the model already has a feature bias for the intended feature (e.g. $h_1$ for NLI and QA) or has low feature bias (Toxicity). They are less effective at overcoming prior feature biases.
    We omit results for steering towards $h_1$ on sentiment analysis since the baseline already has a near-perfect preference for $h_1$ with little room for improvement. 
    Error bars represent the standard deviation over three random seeds, which correspond to three sets of randomly sampled demonstration examples as prompts. 
}    \label{fig:text_davinci_interventions_aggregate}
\end{figure*}

\subsection{Results}

\paragraph{Which interventions are effective?}
Table~\ref{tab:interventions_aggregated} summarizes the results of these experiments, comparing the effect of the different interventions on \textsc{Davinci} and \textsc{Text-Davinci-002}, averaged over features and datasets, and comparing between interventions designed to steer towards the default feature ($h_1$) and the alternative
features.
\emph{Nearly all interventions increase the average $h$-accuracy}, in many cases by a substantial amount.
However, the effectiveness of the intervention varies depending on the model.
For the \textsc{Davinci} model, the best-performing interventions include semantic verbalizers and template-based explanations, while the worst-performing intervention is natural-language instructions.
For the \textsc{Text-Davinci-002} model, instructions are much more effective.

\emph{In some cases, prompt-based interventions are more effective at steering the model than providing unambiguous demonstration examples.} On one hand, this suggests that ICL can be effective even given highly underspecified data, but it also indicates that ICL models may fail to exploit the information provided in the demonstrations.
This finding illustrates that ICL works very differently from standard supervised learning, and calls to mind existing empirical~\citep{min2022rethinking} and theoretical results~\citep{xie2022an} suggesting that ICL might work in part by recognizing existing tasks rather than directly learning the input-output relation from the demonstration.

\paragraph{When are interventions effective?}
Figure~\ref{fig:text_davinci_interventions_aggregate} compares the results of different interventions on the \textsc{Text-Davinci-002} model, aggregated over features. (Detailed results for each feature and \textsc{Davinci} results are in Appendix~\ref{sec:intervention_breakdown}.)
The effectiveness of the intervention varies depending on whether the prior feature bias and the intended feature are aligned.
The interventions are most effective in two scenarios.
First, \emph{interventions are effective when the model already has a feature bias for the intended features}. This is evident in the interventions that steer the model towards $h_1$ in NLI and QA, settings in which the model already has a feature bias in favor of $h_1$.
Second, \emph{interventions are effective when the model has a low feature bias}.
This is the case in the Toxicity Classification dataset, where the model does not exhibit a strong feature bias.
In this setting, all interventions are moderately successful at steering the model towards $h_1$, and more successful at steering the model towards $h_2$.

On the other hand, \emph{interventions are less effective at overriding feature biases}. This trend is illustrated in the second row of Figure~\ref{fig:text_davinci_interventions_aggregate}, in which the intervention is designed to steer the model towards $h_2$ rather than the standard dataset label.
While some interventions increase $h_2$-accuracy, no intervention consistently gets the model to generalize according to the intended feature.

\begin{table*}[t!]
    \centering
    \small
    \setlength{\tabcolsep}{5pt}{
\begin{tabular}{llrrrrr}
\toprule
& $h_2$ & \tf{Govt/Fiction} & \tf{Govt/Telephone} & \tf{Length} & \tf{Negation} & \tf{Overlap} \\
\midrule
\textsc{Davinci} & \ti{Baseline} &         \ti{87.9} &           \ti{89.1} &   \ti{52.3} &     \ti{41.9} &    \ti{69.5} \\
\midrule
 & Verbalizer &          +3.0 &            +1.2 &   -2.0 &     +10.3 &    -6.6 \\
 & Instruction &         -2.2 &          -10.2 &   -1.1 &      +6.1 &    -6.3 \\ 
 & Explanation &          +5.2 &            +6.3 &   -1.2 &      +8.2 &   -19.5 \\
  & Disambig &          +2.3 &            +7.8 &   +14.4 &     +12.7 &    -0.6 \\
\midrule
\textsc{Text-Davinci-002} & \ti{Baseline} &         \ti{28.7} &           \ti{39.5} &   \ti{27.3} &     \ti{34.7} &    \ti{39.2} \\
\midrule
 & Verbalizer &         +66.8 &           +58.2 &   -4.5 &     -0.9 &    -3.2 \\
 & Instruction &         +42.0 &           +41.3 &    +6.8 &     +28.7 &    -3.8 \\
 & Explanation &         +47.7 &           +51.8 &   +26.5 &      +9.1 &     +3.1 \\
 & Disambig &         +37.3 &           +45.9 &   +12.8 &      +0.7 &    +12.7 \\
\bottomrule
\end{tabular}
    }
\caption{
Comparing the effectiveness of intervention strategies on steering the model towards different features in MultiNLI.
This table reports the $h_2$-accuracy in the baseline setting for each feature and the difference in $h_2$-accuracy obtained by the different interventions.
The interventions are generally more successful on simple semantic features like genre and less successful on semantically irrelevant features like lexical overlap and length.
}
\label{tab:interventions_features}
\end{table*}

\paragraph{Which features are most susceptible to interventions?}
In Table~\ref{tab:interventions_features}, we compare the effect of interventions on different features in MultiNLI.
Using meaningful label words works better on the genre features, where the label words are semantically similar to the input example, but it is more difficult to steer the model toward the use of features like length and lexical overlap, which are not related to the semantics of the sentences.
More work may be needed to develop interventions that work well for higher-order language features.

Lastly, we compile a \emph{summary of practical takeaways} for steering LLMs:
\begin{itemize}
    \item When using non-instruction-tuned LLMs (e.g., \textsc{Davinci}), specifying feature preferences as instructions is not effective, instead adding explanations or disambiguating examples can be more effective. 

    \item When using instruction-tuned LLMs (e.g., \textsc{Text-Davinci-002}), all interventions can be effective. 

    \item Features not related to semantics, such as sentence lengths or overlap, are difficult to intervene across all conditions. 
\end{itemize}


\section{Related Work}


\paragraph{Measuring inductive biases.} Our work builds on existing research on measuring the inductive biases of learning algorithms in machine learning and NLP.
~\citet{dasgupta2022distinguishing} introduce a methodology for measuring feature bias as well as rule- vs. exemplar-based generalization,
and~\citet{chan2022transformers} apply this approach to compare rule- vs. exemplar-based learning in ICL.
We use a similar framing as~\citet{dasgupta2022distinguishing}, but focus on feature bias.
In NLP, another line of work has studied the inductive biases of neural networks in the context of the poverty of the stimulus argument~\citep{chomsky1980rules}.
These studies evaluate whether neural networks generalize in a manner consistent with structural or superficial linguistic rules~\citep{mccoy2018revisiting,mccoy2020does}.
Several studies have found that models such as BERT acquire a preference for structural generalizations from large-scale masked language model pretraining~\citep{Warstadt2020CanNN,warstadt2020learning,zhang2021inductive,mueller-etal-2022-coloring}.
We follow a similar poverty-of-the-stimulus experimental setup but focus on in-context learning and on features arising in common NLP tasks.
~\citet{Lovering2021PredictingIB} explore whether it is possible to predict the inductive biases of pre-trained models and show that models tend to generalize on the basis of features that are more ``extractable'', measured using probing techniques~\citep{voita2020information}, but it is not straightforward to extend the notion of extrability to in-context learning.
~\citet{tamkin2022task} also study how LLMs generalize to ambiguous classification tasks, but focus on ambiguous instructions and use template-generated datasets.


\paragraph{Spurious correlations.}
A related line of work has explored the inductive biases of pretrained LMs in relation to spurious correlations, or shortcuts \citep[e.g. ][]{Gururangan2018AnnotationAI,poliak2018hypothesis,HANS,geirhos2020shortcut,sagawa2020investigation}---shallow features that are correlated with the classification targets.
Models can generalize successfully if they have an inductive bias that tends to favor the intended feature over the shortcut. ~\citet{pmlr-v97-hendrycks19a,TransformerOOD,Tu2020AnES} found that pre-trained models can generalize more successfully to such distribution shifts.



\paragraph{Explaining in-context learning.}
A number of empirical studies have attempted to characterize the behavior of ICL and explain why it works.
These studies have found that ICL can be overly sensitive to certain aspects of the prompt, such as the order of the demonstration examples~\citep{lu2022fantastically} and choice of label words~\citep{zhao2021calibrate}, but also surprisingly insensitive to others.
In particular, \citet{min2022rethinking} show that LLMs can largely ignore the relationship between inputs and labels in the demonstration example
and~\citet{webson2022prompt} show that the performance of ICL can perform well on NLI even if the prompt is irrelevant or misleading.
Relatedly, \citet{Wei2023LargerLM,pan2023what} show that LLMs can perform ICL well with flipped labels or semantically-unrelated labels, but such ability of overriding semantic priors emerges with scale.
Some theoretical work has also attempted to explain why prompt-based methods work, by drawing connections between the prompting setting and properties of the pretraining distribution~\citep{saunshi2021a,wei2021pretrained,xie2022an} or by arguing that Transformers can act as meta-learners, implicitly performing gradient descent on the in-context examples~\citep{von2022transformers,akyurek2022learning}. Our results provide empirical evidence that there is a strong task bias from pretraining when the LLMs must infer the task by input-output relations.

\paragraph{Improving in-context learning.} Recent work studied the effect of including explanations in the prompt to produce better quality answers~\cite{Wei2022ChainOT,Lampinen2022CanLM}. While they show the benefit of high-quality human-annotated explanations for improving task performance, we demonstrated the effectiveness of simple template explanations in steering feature biases. Another line of work collects large pools of instructions from diverse tasks and uses them to tune and control language models~\cite{Wang2022SuperNaturalInstructionsGV,Chung2022ScalingIL}. We also adopt instructions as an intervention method and show that it works particularly well on instruction-tuned models. In a similar manner, \citet{Si2022PromptingGT} studied prompting methods to make GPT-3 more reliable, such as instructing it to not rely on demographic biases. 

\section{Conclusion}

In this work, we constructed underspecified prompts from real-world datasets to study feature biases of large language models. We found that the instruction-tuned InstructGPT model prefers the ``default'' task features over distractor features more often than the base GPT-3 model, and we demonstrated the effectiveness of various intervention methods in steering models to use the specified feature. These results not only shed new insights into the working mechanisms of ICL, but also have practical takeaways for discouraging models from exploiting unintended features such as demographic biases or shallow statistical cues.

\section*{Limitations}

\paragraph{Model coverage.} Our study is targeted specifically at GPT-3 and it would be interesting to study feature bias patterns on other large language models such as OPT~\cite{Zhang2022OPTOP} and BLOOM~\cite{Scao2022BLOOMA1}; and it is possible that our intervention methods may have different effects on these language models trained with different data sources and scales. 

\paragraph{Task coverage.} Apart from model coverage, our analysis is focused on only four common binary classification tasks.
Our main metric, $h$-accuracy, compares the predictions between a learned function $f$ and a feature function $h$. For simplicity, we only study binary functions (consistent with prior work) to illustrate the main ideas, but the framework applies equally well if $f$ and $h$ are multi-class classifiers. For example, in the case of the three-way NLI task, we might set $h_1$ to predict on the basis of entailment / contradiction / neutral, and $h_2$ to predict on the basis of the genres – e.g. fiction / government / telephone. Future work could extend our framework to more tasks, and consider how to apply it to more complex tasks such as generation. 

\paragraph{Feature coverage.} Our current experiments are limited to a set of hand-crafted features. One potential way to systematically scale our approach is to develop novel techniques for automatic feature discovery, for example, to cluster the data and treat each cluster as having a distinct feature. 


\paragraph{Explaining feature biases.} While our empirical findings shed light on the feature bias patterns of GPT-3, we do not yet have a conclusion on how these biases arise from pretraining. Future work could attempt to draw connections to the pretraining data or to theoretical accounts of in-context learning.  

\paragraph{Risks and ethics.}
While we do not foresee any ethical risks resulting from our work, we caution against making extrapolations about the extent to which LLMs exhibit feature biases towards protected social attributes.
Although we do not find evidence of strong feature biases in a particular toxicity classification setting, care should be taken to evaluate the fairness and reliability of these models directly before they are deployed in downstream applications.

\section*{Acknowledgement}

We thank Alex Tamkin, Xi Ye, Sewon Min, and Jordan Boyd-Graber for their helpful feedback. This research is partially funded by the National Science Foundation (IIS-2211779), a Sloan research fellowship, Samsung Advanced Institute of Tech-
nology (under the project Next Generation Deep
Learning: From Pattern Recognition to AI) and
AWS AI. NJ is supported by an NSF Graduate Research Fellowship under grant number 1839302.

\bibliography{ref}
\bibliographystyle{acl_natbib}

\clearpage 
\appendix
\section{Appendix}


\subsection{Intervention Results Across Features}
\label{sec:intervention_breakdown}

We present the complete set of intervention results broken down by features in Table~\ref{tab:intervention-Davinci} for \textsc{Davinci} and Table~\ref{tab:intervention-TD002} for \textsc{Text-Davinci-002}. 

It is worth noting that the intervention methods' effectiveness often varies across features even on the same dataset. For example, all intervention methods can effectively steer models towards using the genre feature over the entailment feature on MNLI, but the success is  limited for the lexical overlap feature on MNLI. We hypothesize this is because features like lexical overlap are harder for models to recognize.

\begin{table*}[ht!]
    \centering
    \small
    \setlength{\tabcolsep}{5pt}{
    \begin{tabular}{ lcccccc }
    \toprule 
     & $h_2$  
     & Baseline
     & + DisAmbig 
     & + Verbalizers
     & + Instruction 
     & + Explanation \\
     \midrule 
    \multicolumn{7}{c}{\textit{Steer Towards $h_1$}} \\
    \midrule 
    NLI & genre & 11.5 / 88.5 & 53.4 / 46.6 & 28.5 / 71.5 & 13.5 / 86.5 & 20.9 / 79.1 \\
    &  length & 47.7 / 52.3 & 54.5 / 45.5 & 53.3 / 46.7 & 49.3 / 50.7 & 53.4 / 46.6 \\
    & negation & 58.1 / 41.9 & 49.7 / 50.3 & 52.6 / 47.4 & 51.8 / 48.2 & 45.4 / 54.6 \\
    & overlap & 30.5 / 69.5 & 54.9 / 45.1 & 29.4 / 70.6 & 37.9 / 62.1 & 44.5 / 55.5 \\
    & Aggregate & 31.9 / 68.1 & 53.2 / 46.8 & 38.4 / 61.6 & 33.2 / 66.8 & 37.0 / 63.0 \\
    \midrule 
    TC & gender & 47.9 / 52.1 & 50.2 / 49.8 & 53.1 / 46.9 & 47.8 / 52.2 & 58.6 / 41.4 \\
    & race & 53.0 / 47.0 & 51.1 / 48.9 & 58.0 / 42.0 & 52.2 / 47.8 & 59.5 / 40.5 \\
    & religion & 51.4 / 48.6 & 50.7 / 49.3 & 53.4 / 46.6 & 51.9 / 48.1 & 55.1 / 44.9 \\
     & length & 36.6 / 63.4 & 53.6 / 46.4 & 33.8 / 66.2 & 37.8 / 62.2 & 62.6 / 37.4 \\
     & capitalization & 32.4 / 67.6 & 51.2 / 48.8 & 51.5 / 48.5 & 30.0 / 70.0 & 60.3 / 39.7 \\
      & Aggregate & 45.8 / 54.2 & 51.1 / 48.9 & 50.9 / 49.1 &  45.6 / 54.4 & 58.6 / 41.4 \\
    \midrule 
    QA & Q word & 39.8 / 60.2 & 63.5 / 36.5 & 76.1 / 23.9 & 45.1 / 54.9 & 67.2 / 32.8 \\
    & overlap & 41.1 / 58.9 & 55.1 / 44.9 & 73.1 / 26.9 & 44.9 / 55.1 & 64.3 / 35.7  \\
    & structure & 34.0 / 66.0 & 49.3 / 50.7 & 61.7 / 38.3 & 37.4 / 62.6 & 71.0 / 29.0 \\
    & length & 37.8 / 62.2 & 48.4 / 51.6 & 64.2 / 35.8  & 46.2 / 53.8 & 64.8 / 35.2 \\
    & Aggregate & 38.2 / 61.8 & 54.1 / 45.9 & 68.8 / 31.2  & 43.4 / 56.6 & 66.8 / 33.2 \\
    \midrule
    \multicolumn{7}{c}{\textit{Steer Towards $h_2$}} \\
    \midrule
   NLI & genre & 11.5 / 88.5 & 6.5 / 93.5 & 9.4 / 90.6 & 17.7 / 82.3 & 5.7 / 94.3 \\
    &  length & 47.7 / 52.3 & 33.3 / 66.7 & 49.8 / 50.2 & 48.8 / 51.2 & 48.8 / 51.2 \\
    & negation & 58.1 / 41.9 & 45.4 / 54.6 & 47.8 / 52.2 & 52.0 / 48.0 & 49.8 / 50.2 \\
    & overlap & 30.5 / 69.5 & 31.1 / 68.9 & 37.1 / 62.9 & 36.7 / 63.3 & 50.0 / 50.0 \\
    & Aggregate & 31.9 / 68.1 & 24.6 / 75.4 & 30.7 / 69.3 & 34.5 / 65.5 & 32.0 / 68.0 \\
    \midrule 
    SA & punctuation & 98.3 / 1.7 & 73.1 / 26.9 & 97.0 / 3.0 & 98.0 / 2.0 & 68.6 / 31.4 \\
    & domain & 56.1 / 43.9 & 0.3 / 99.7 & 1.0 / 99.0 & 77.6 / 22.4 & 25.8 / 74.2  \\
    & length & 98.4 / 1.6 & 35.2 / 64.8 & 30.7 / 69.3 & 97.8 / 2.2 & 62.1 / 37.9 \\
    & lexicon & 95.5 / 4.5 & 63.2 / 36.8 & 87.8 / 12.2 & 96.5 / 3.5 & 72.0 / 28.0  \\
    & capitalization & 92.0 / 8.0 & 43.5 / 56.5 & 85.5 / 14.5 & 81.5 / 18.5 & 75.2 / 24.8 \\
     & Aggregate & 89.3 / 10.7 & 46.4 / 53.6 & 65.0 / 35.0 & 91.3 / 8.7 & 62.6 / 37.4 \\
    \midrule 
    TC & gender & 47.9 / 52.1  & 41.3 / 58.7 & 29.8 / 70.2 & 48.3 / 51.7 & 28.8 / 71.2 \\
    & race & 53.0 / 47.0 &  38.4 / 61.6 & 26.2 / 73.8 & 50.9 / 49.1 & 27.1 / 72.9 \\
    & religion & 51.4 / 48.6 & 34.6 / 65.4 & 16.8 / 83.2 & 51.1 / 48.9 & 9.8 / 90.2 \\
     & length & 36.6 / 63.4 &  29.7 / 70.3 & 22.7 / 77.3 & 40.4 / 59.6 & 31.2 / 68.8 \\
     & capitalization & 32.4 / 67.6 & 18.1 / 81.9 & 48.8 / 51.2 & 31.6 / 68.4 & 43.0 / 57.0 \\
      & Aggregate & 45.8 / 54.2 & 34.0 / 66.0 & 27.3 / 72.7 & 46.1 / 53.9 & 25.5 / 74.5 \\
    \midrule 
    QA & Q word & 39.8 / 60.2 & 31.1 / 68.9 & 11.5 / 88.5 & 43.1 / 56.9 & 1.7 / 98.3 \\
    & overlap & 41.1 / 58.9 & 28.9 / 71.1 & 50.0 / 50.0 & 45.5 / 54.5  & 49.9 / 50.1 \\
    & structure & 34.0 / 66.0 & 16.3 / 83.7 & 55.5 / 44.5 & 39.3 / 60.7 & 48.3 / 51.7 \\
    & length & 37.8 / 62.2 & 43.4 / 56.6 & 37.0 / 63.0 & 47.7 / 52.3 & 34.3 / 65.7 \\
    & Aggregate & 38.2 / 61.8 & 29.9 / 70.1 & 38.5 / 61.5 & 43.9 / 56.1 & 33.6 / 66.4 \\
    \bottomrule
    \end{tabular}
    }
    \caption{The impact of different intervention strategies (applied separately on top of the baseline). This table is for \textsc{Davinci}. We report the ambiguous accuracy for supporting $h_1$ and $h_2$ respectively in each cell, higher $h_1$ accuracy indicates a preference for the $h_1$ hypothesis. Most interventions successfully steer the model preference in the intended direction. We omit results for steering towards $h_1$ on sentiment analysis since both models already have a strong preference for the sentiment feature without any intervention. 
    We average and condense results of features from the same category (such as gender and sexuality on CivilComments, and different lexicon features on sentiment analysis) since their trends are largely similar. 
    }
    \label{tab:intervention-Davinci}
    \end{table*}

\begin{table*}[ht!]
    \centering
    \small
    \setlength{\tabcolsep}{5pt}{
    \begin{tabular}{ lcccccc }
    \toprule 
     & h$_2$  
     & Baseline
     & + DisAmbig 
     & + Verbalizers
     & + Instruction 
     & + Explanation \\
     \midrule 
    \multicolumn{7}{c}{\textit{Steer Towards $h_1$}} \\
    \midrule 
    NLI & genre & 65.9 / 34.1 & 79.2 / 20.8 & 74.9 / 25.1 & 86.1 / 13.9 & 75.9 / 24.1 \\
    &  length & 72.7 / 27.3 & 83.6 / 16.4 & 73.4 / 26.6 & 81.6 / 18.4 & 77.8 / 22.2 \\
    & negation & 65.3 / 34.7 & 72.5 / 27.5  & 77.3 / 22.7 & 80.1 / 19.9 & 77.8 / 22.2 \\
    & overlap & 60.8 / 39.2 & 73.1 / 26.9 & 64.6 / 35.4 & 70.9 / 29.1 & 66.4 / 33.6 \\
    & Aggregate & 66.1 / 33.9 & 77.5 / 22.5 & 73.0 / 27.0 & 81.0 / 19.0 & 74.8 / 25.2 \\
    \midrule 
    TC & gender & 45.9 / 54.1 & 55.2 / 44.8 & 53.0 / 47.0 & 58.4 / 41.6 & 60.4 / 39.6 \\
    & race & 50.9 / 49.1 & 59.7 / 40.3 & 55.8 / 44.2 & 60.8 / 39.2 & 60.7 / 39.3 \\
    & religion & 42.0 / 58.0 & 56.8 / 43.2 & 49.5 / 50.5 & 56.5 / 43.5 & 57.7 / 42.3 \\
     & length & 45.9 / 54.1 & 60.8 / 39.2 & 53.3 / 46.7  & 66.8 / 33.2 & 62.7 / 37.3 \\
     & capitalization & 48.8 / 51.2 & 58.1 / 41.9 & 55.3 / 44.7 & 63.8 / 36.2 & 64.9 / 35.1 \\
      & Aggregate & 45.9 / 54.1 & 57.5 / 42.5 & 52.8 / 47.2 & 60.1 / 39.9 & 60.6 / 39.4 \\
    \midrule 
    QA & Q word & 79.7 / 20.3 & 77.2 / 22.8 & 85.2 / 14.8 & 84.3 / 15.7 & 86.1 / 13.9 \\
    & overlap & 71.3 / 28.7 & 77.6 / 22.4 & 81.1 / 18.9 & 76.7 / 23.3 & 81.6 / 18.4  \\
    & structure & 68.0 / 32.0 & 72.3 / 27.7 & 78.8 / 21.2 & 71.6 / 28.4 & 74.9 / 25.1 \\
    & length & 74.6 / 25.4 & 78.7 / 21.3 & 80.4 / 19.6 & 82.6 / 17.4 & 85.7 / 14.3 \\
    & Aggregate & 73.4 / 26.6 & 76.5 / 23.5 & 81.4 / 18.6 & 78.8 / 21.2 & 82.0 / 18.0 \\
    \midrule
    \multicolumn{7}{c}{\textit{Steer Towards $h_2$}} \\
    \midrule
   NLI & genre & 65.9 / 34.1 & 24.3 / 75.7 & 3.4 / 96.6 & 24.2 / 75.8 & 16.1 / 83.9 \\
    &  length & 72.7 / 27.3 & 59.9 / 40.1  & 77.2 / 22.8 & 65.9 / 34.1 & 46.2 / 53.8 \\
    & negation & 65.3 / 34.7 & 64.6 / 35.4 & 66.2 / 33.8 & 36.6 / 63.4 & 56.2 / 43.8 \\
    & overlap & 60.8 / 39.2 & 48.1 / 51.9 & 63.9 / 36.1 & 64.6 / 35.4 & 57.7 / 42.3 \\
    & Aggregate & 66.1 / 33.9 & 44.2 / 55.8 & 42.8 / 57.2 & 43.1 / 56.9 & 38.4 / 61.5 \\
    \midrule 
    SA & punctuation & 99.1 / 0.9 & 98.0 / 2.0 & 96.6 / 3.4 & 85.4 / 15.6 & 50.5 / 49.5 \\
    & domain & 92.4 / 7.6 & 0.7 / 99.3 & 0.5 / 99.5 & 1.1 / 98.9 & 26.4 / 73.6 \\
    & length & 98.6 / 1.4  & 76.1 / 23.9 & 27.7 / 72.3 & 81.1 / 18.9 & 42.6 / 57.4 \\
    & lexicon & 97.8 / 2.2 & 65.7 / 34.3 & 87.6 / 22.4 & 68.6 / 31.4 & 67.4 / 32.6 \\
    & capitalization & 98.4 / 1.6 & 56.2 / 43.8 & 87.2 / 12.8 & 61.7 / 38.3 & 56.0 / 44.0 \\
     & Aggregate & 97.3 / 2.7 & 60.4 / 39.6 & 64.5 / 35.5 & 61.1 / 38.9 & 51.7 / 48.3 \\
    \midrule 
    TC & gender & 45.9 / 54.1 & 37.9 / 62.1 & 41.0 / 59.0 & 22.9 / 77.1 & 12.1 / 87.9 \\
    & race & 50.9 / 49.1 & 42.7 / 57.3 & 36.0 / 64.0 & 33.1 / 66.9 & 7.4 / 92.6 \\
    & religion & 42.0 / 58.0 & 21.8 / 78.2 & 7.7 / 92.3 & 9.4 / 90.6 & 5.7 / 94.3 \\
     & length & 45.9 / 54.1 & 31.0 / 69.0 & 20.3 / 79.7 & 19.6 / 80.4 & 3.6 / 96.4 \\
     & capitalization & 48.8 / 51.2 & 28.9 / 71.1 & 41.3 / 58.7 & 43.7 / 56.3 & 48.0 / 52.0 \\
      & Aggregate & 45.9 / 54.1 & 31.7 / 68.3 & 27.8 / 72.2 & 23.0 / 77.0 & 13.5 / 86.5 \\
    \midrule 
    QA & Q word & 79.7 / 20.3 & 68.9 / 31.1 & 0.1 / 99.9 & 83.5 / 16.5 & 39.4 / 60.6 \\
    & overlap & 71.3 / 28.7 & 66.1 / 33.9 & 78.5 / 21.5 & 69.3 / 30.7 & 64.5 / 35.5 \\
    & structure & 68.0 / 32.0  & 62.0 / 38.0 & 62.5 / 37.5 & 48.9 / 51.1 & 33.9 / 66.1 \\
    & length & 74.6 / 25.4 & 62.1 / 37.9 & 36.0 / 64.0 & 51.5 / 48.5 & 49.9 / 50.1  \\
    & Aggregate & 73.4 / 26.6 & 64.8 / 35.2 & 44.3 / 55.7 & 63.3 / 36.7 & 46.9 / 53.1 \\
    \bottomrule
    \end{tabular}
    }
    \caption{The impact of different intervention strategies (applied separately on top of the baseline). This table is for \textsc{Text-Davinci}-002. We report the ambiguous accuracy for supporting $h_1$ and $h_2$ respectively in each cell, higher $h_1$ accuracy indicates a preference for the $h_1$ hypothesis. Most interventions successfully steer the model preference in the intended direction. We omit results for steering towards $h_1$ on sentiment analysis since both models already have a strong preference for the sentiment feature without any intervention. 
    We average and condense results of features from the same category (such as gender and sexuality on CivilComments, and different lexicon features on sentiment analysis) since their trends are largely similar. 
    }
    \label{tab:intervention-TD002}
    \end{table*}

\subsection{List of Semantic Verbalizers}
\label{sec:verbalizers}

\begin{table*}[ht!]
\small
\begin{center}
\begin{tabular}{l l l}
\toprule
\textbf{Task} & \textbf{Feature} & \textbf{Verbalizer} \\
\midrule
Sentiment analysis & Sentiment & 1- ``positive'', 0 - ``negative'' \\
& Domain & ``Source?'' 1 - ``movie'', 0 - ``other'' \\
& Length & ``Length?'' 1 - ``short'', 0 - ``long'' \\
& Terminal punctuation & ``End punctuation?'' 1 - ``other'', 0 - ``period'' \\
& Contains word  & ``Has \textit{food / nice}?'' 1 - ``yes'', 0 - ``no'' \\
& Capitalization & ``Uppercase words?'' 1 - ``yes'', 0 - ``no'' \\
Toxicity classification &  Toxicity & ``Toxic?'' 1 - ``yes'', 0 - ``no'' \\
&  Gender  & ``Gender?'' 1 - ``female'', 0 - ``male'' \\
& Sexuality & ``LGBTQ?'' 1 - ``yes'', 0 - ``no''  \\
& Religion & ``Religion?'' 1 - ``Muslim'', 0 - ``Christian'' / ``Jewish'' \\
&  Race & ``Race?'' 1 - ``black'', 0 - ``white'' \\
&  Length & ``Length?'' 1 - ``short'', 0 - ``long'' \\
&  Capitalization & ``Uppercase words?'' 1 - ``yes'', 0 - ``no'' \\
Natural language inference  & Entailment & ``Entailed?'' 1 - ``yes'', 0 - ``no'' \\
&  Domain  & ``Source?'' 1 - ``government'', 0 - ``fiction'' / ``telephone'' \\
&  Lexical overlap & ``Overlap?'' 1 - ``yes'', 0 - ``no'' \\
&  Hypothesis length  & ``Shorter?'' 1 - ``yes'', 0 - ``no'' \\
&  Hypothesis negation  & ``Negation?'' 1 - ``yes'', 0 - ``no'' \\
Question answering  & Answer & 1 - ``yes'', 0 - ``no'' \\
& Question word & ``Question word?'' 1 - ``be'', 0 - ``do'' \\
& Lexical overlap & ``Overlap?'' 1 - ``yes'', 0 - ``no'' \\
& Question structure & ``Has \textit{same as}?'' 1 - ``yes'', 0 - ``no'' \\
& Passage length & ``Length?'' 1 - ``long'', 0 - ``short'' \\
\bottomrule
\end{tabular}
 \caption{All verbalizers we used in our experiments. For some hypotheses, it is difficult to use succinct words to specify the label (e.g., whether it contains negation words), we thus prepend a short question before the label words for such cases to better specify the task (e.g., ``Negation? yes / no'').}
 \label{tab:verbalizers}
\end{center}
\end{table*}

We present the full list of semantically meaningful verbalizers for the intervention experiments in Table~\ref{tab:verbalizers}.  

\subsection{List of Task Instructions}
\label{sec:instructions}

\begin{table*}[ht!]
\small
\begin{center}
\begin{tabularx}{\textwidth}{l l X}
\toprule
\textbf{Task} & \textbf{Feature} & \textbf{Instruction} \\
\midrule
Sentiment Analysis & Sentiment & Given a review, you need to predict whether the review is good (positive) or bad (negative). Please answer with ``1'' for positive and ``0'' for negative. \\
& Domain & Given a review, you need to predict whether it comes from IMDB movie reviews or Yelp restaurant and service reviews. Please answer with ``1'' for IMDB and ``0'' for Yelp. \\
& Length & Given a review, you need to predict whether it is longer than 40 words. Please answer with ``1'' for shorter than 40 words and ``0'' for longer. \\
& Terminal punctuation & Given a review, you need to predict whether the review ends with an exclamation mark (`!') or period (`.'). Please answer with ``1'' for exclamation mark and ``0'' for period. \\
& Contains word  & Given a review, you need to predict whether the review mentions the word ``nice''/``food''. Please answer with ``1'' for mentioning ``nice''/``food'' and ``0'' for not mentioning. \\
& Capitalization &  Given a review, you need to predict whether the review contains any uppercase words where all letters are uppercase (like `THIS'). Please answer with ``1'' for having uppercase words and ``0'' for not. \\
Toxicity classification &  Toxicity & Categorize the comment on the basis of toxicity. If the comment is toxic (harmful, malicious, derogatory, threat, insult, identity attack, etc.) output ``1'', otherwise output ``0''. \\
&  Gender  & Given a comment, you need to predict whether the comment targets males or females. Please answer with ``1'' for female and ``0'' for male. \\
& Sexuality &  Given a comment, you need to predict whether the comment targets LGBTQ people. Please answer with ``1'' if it does and ``0'' if not. \\
& Religion &  Given a comment, you need to predict whether the comment targets Muslim or Christian/Jewish people. Please answer with ``1'' for Muslim and ``0'' for Christian/Jewish. \\
&  Race &  Given a comment, you need to predict whether the comment targets black or white people. Please answer with ``1'' for black people and ``0'' for white people. \\
&  Length &  Given a comment, you need to predict whether the comment is longer than 40 words. Please answer with ``1'' for shorter and ``0'' for longer. \\
&  Capitalization & Given a comment, you need to predict whether the comment contains any uppercase words where all letters are uppercased (like `THIS'). Please answer with ``1'' for having uppercase words and ``0'' for not. \\
\bottomrule
\end{tabularx}
 \caption{All instructions used in our experiments. We prepend the corresponding instruction of each hypothesis to the prompt. This table is for sentiment analysis and toxicity classification.}
 \label{tab:instructions_1}
\end{center}
\end{table*}

\begin{table*}[ht!]
\small
\begin{center}
\begin{tabularx}{\textwidth}{l l X}
\toprule
\textbf{Task} & \textbf{Feature} & \textbf{Instruction} \\
\midrule
Natural language inference  & Entailment & In this task, you will be presented with a premise sentence (the first sentence) and a hypothesis sentence (the second sentence). Determine whether the premise sentence entails (implies) or does not entail the hypothesis sentence. Please answer with ``1'' for entailment and ``0'' for non-entailment. \\
&  Domain  & In this task, you will be presented with a premise sentence (the first sentence) and a hypothesis sentence (the second sentence). Determine whether they come from government files or fiction/telephone. Please answer with ``1'' for government and ``0'' for fiction \\
&  Lexical overlap & In this task, you will be presented with a premise sentence (the first sentence) and a hypothesis sentence (the second sentence). Determine whether all words in the second sentence also appear in the first sentence. If so, answer ``1''; if not, answer ``0''. \\
&  Hypothesis length  &  In this task, you will be presented with a premise sentence (the first sentence) and a hypothesis sentence (the second sentence). Determine whether the second sentence is shorter than the first sentence. Please answer with ``1'' for shorter and ``0'' for longer. \\
&  Hypothesis negation  &  In this task, you will be presented with a premise sentence (the first sentence) and a hypothesis sentence (the second sentence). Determine whether there are any negation words in the second sentence (``not'', ``no'', ``n't''). Please answer with ``1'' for not having negations and ``0'' for having negations. \\
Question answering  & Answer & Based on the information present in the given passage, decide whether the answer to the given question is yes or no. Please answer with ``1'' for yes and ``0'' for no. \\
& Question word &  Given the passage and question, determine whether the question word is ``is/was'' or ``do/does/did''. Please answer with ``1'' for ``is/was'' and ``0'' for ``do/does/did''. \\
& Lexical overlap & Given the passage and question, determine whether all words in the question also appear in the passage. If so, answer ``1''; if not, answer ``0''.  \\
& Question structure & Given the passage and question, determine whether the question contains the phrase ``same as''. Please answer with ``1'' for having ``same as'' and ``0'' if not. \\
& Passage length & Given the passage and question, determine whether the passage is longer than 50 words. Please answer with ``1'' for longer than 50 words and ``0'' for shorter. \\
\bottomrule
\end{tabularx}
 \caption{All instructions used in our experiments. We prepend the corresponding instruction of each hypothesis to the prompt. This table is for natural language inference and question answering.}
 \label{tab:instructions_2}
\end{center}
\end{table*}

We present the full list of task instructions for the intervention experiments in Table~\ref{tab:instructions_1} and Table~\ref{tab:instructions_2}. 

\subsection{List of Template Explanations}
\label{sec:explanations}

\begin{table*}[ht!]
\small
\begin{center}
\begin{tabularx}{\textwidth}{l l X}
\toprule
\textbf{Task} & \textbf{Feature} & \textbf{Explanation} \\
\midrule
Sentiment Analysis & Sentiment (1) & The review has a positive sentiment. Therefore, the answer is 1. \\
 & Sentiment (0) & The review has a negative sentiment. Therefore, the answer is 0. \\
& Domain (1) & The review is from IMDB movie reviews. Therefore, the answer is 1. \\
& Domain (0) & The review is from Yelp reviews. Therefore, the answer is 0. \\
& Length (1) &  The review is shorter than 40 words. Therefore, the answer is 1. \\
& Length (0) & The review is longer than 40 words. Therefore, the answer is 0. \\
& Terminal punctuation (1) &  The review ends with an exclamation mark (`!'). Therefore, the answer is 1. \\
& Terminal punctuation (0) & The review ends with a period ('.'). Therefore, the answer is 0. \\
& Contains word (1) & The review contains the word `food'/`nice'. Therefore, the answer is 1. \\
& Contains word (0) &  The review does not contain the word `food'/`nice'. Therefore, the answer is 0. \\
& Capitalization (1) &  The review contains an uppercase word with all uppercase letters. Therefore, the answer is 1. \\
& Capitalization (0) & The review does not contain an uppercase word with all uppercase letters. Therefore, the answer is 0. \\
Toxicity classification &  Toxicity (1) & The comment is toxic. Therefore, the answer is 1.  \\
&  Toxicity (0) & The comment is not toxic. Therefore, the answer is 0. \\
&  Gender (1) & The comment mentions females. Therefore, the answer is 1. \\
&  Gender (0) &  The comment mentions males. Therefore, the answer is 0. \\
& Sexuality (1) &  The comment mentions LGBTQ. Therefore, the answer is 1. \\
& Sexuality (0) &  The comment does not mention LGBTQ. Therefore, the answer is 0. \\
& Religion (1) &  The comment mentions Muslim people. Therefore, the answer is 1. \\
& Religion (0) &  The comment mentions Christian/Jewish people. Therefore, the answer is 0. \\
&  Race (1) &  The comment mentions black people. Therefore, the answer is 1. \\
&  Race (0) &  The comment mentions white people. Therefore, the answer is 0. \\
&  Length (1) &  The comment is shorter than 40 words. Therefore, the answer is 1. \\
&  Length (0) &  The comment is longer than 40 words. Therefore, the answer is 0. \\
&  Capitalization (1) & The comment contains an uppercase word with all uppercase letters. Therefore, the answer is 1. \\
&  Capitalization (0) & The comment contains an uppercase word with all uppercase letters. Therefore, the answer is 0. \\
\bottomrule
\end{tabularx}
 \caption{All template explanations used in our experiments. The explanation is appended after each input text and before the label for all demonstration examples. Such explanation would be induced during test inference as well. We manually write a template explanation for each class of each hypothesis. This table is for sentiment analysis and toxicity classification.}
 \label{tab:explanations_1}
\end{center}
\end{table*}

\begin{table*}[ht!]
\small
\begin{center}
\begin{tabularx}{\textwidth}{l l X}
\toprule
\textbf{Task} & \textbf{Feature} & \textbf{Explanation} \\
\midrule
Natural language inference  & Entailment (1) & The first sentence entails the second sentence. Therefore, the answer is 1. \\
& Entailment (0) & The first sentence does not entail the second sentence. Therefore, the answer is 0. \\
&  Domain (1) & The text is from government files. Therefore, the answer is 1. \\
&  Domain (0) & The text is from fiction / telephone recordings. Therefore, the answer is 0. \\
&  Lexical overlap (1) & All words from the second sentence also appear in the first sentence. Therefore, the answer is 1. \\
&  Lexical overlap (0) & Not all words from the second sentence also appear in the first sentence. Therefore, the answer is 0. \\
&  Hypothesis length (1)  & The second sentence is shorter than the first sentence. Therefore, the answer is 1. \\
&  Hypothesis length (0)  & The second sentence is longer than the first sentence. Therefore, the answer is 0. \\
&  Hypothesis negation  (1) & The second sentence contains negation words. Therefore, the answer is 1. \\
&  Hypothesis negation  (0) & The second sentence does not contain negation words. Therefore, the answer is 0. \\
Question answering  & Answer (1) &  The answer to the question is yes. Therefore, the answer is 1. \\
 & Answer (0) & The answer to the question is no. Therefore, the answer is 0. \\
& Question word (1) & The question word is `is' or `was'. Therefore, the answer is 1.  \\
& Question word (0) &  The question word is `do' or `does' or `did'. Therefore, the answer is 0. \\
& Lexical overlap (1) &  All words from the question also appear in the passage. Therefore, the answer is 1. \\
& Lexical overlap (0) &  Not all words from the question also appear in the passage. Therefore, the answer is 0. \\
& Question structure (1) & The question contains the phrase `same as'. Therefore, the answer is 1. \\
& Question structure (0) & The question does not contain the phrase `same as'. Therefore, the answer is 0. \\
& Passage length (1) & The passage is longer than 50 words. Therefore, the answer is 1. \\
& Passage length (0) & The passage is shorter than 50 words. Therefore, the answer is 0. \\
\bottomrule
\end{tabularx}
 \caption{All template explanations used in our experiments. The explanation is appended after each input text and before the label for all demonstration examples. Such explanation would be induced during test inference as well. We manually write a template explanation for each class of each hypothesis. This table is for natural language inference and question answering.}
 \label{tab:explanations_2}
\end{center}
\end{table*}

We present the full list of template explanations for the intervention experiments in Table~\ref{tab:explanations_1} and Table~\ref{tab:explanations_2}.

\end{document}